\documentclass[letterpaper, 10 pt, conference]{IEEEtran} 

\IEEEoverridecommandlockouts

\usepackage{cite}
\usepackage{amsmath,amssymb,amsfonts}
\usepackage{algorithmic}
\usepackage{graphicx}
\usepackage{textcomp}
\usepackage{xcolor}
\def\BibTeX{{\rm B\kern-.05em{\sc i\kern-.025em b}\kern-.08em
    T\kern-.1667em\lower.7ex\hbox{E}\kern-.125emX}}

\usepackage{pgfplots}
\usepgfplotslibrary{groupplots}
\usetikzlibrary{calc}
\pgfplotsset{compat=1.18}
\usepackage{dsfont}

\usepackage{comment}
\usepackage{float}
\usepackage{graphicx}
\usepackage{hyperref}
\usepackage{tikz}
\usepackage{pifont}      
\usepackage{listings}
\usepackage{booktabs}

\usepackage{multirow}
\usepackage{pgfplots}
\usepackage{pgfplotstable}
\usepackage{tabularx}
\usepackage{bbm}

\definecolor{dolphinsColor}{HTML}{2E86AB}
\definecolor{omnidriveColor}{HTML}{7B2CBF}
\definecolor{leapvadColor}{HTML}{F77F00}

\newcommand{\dolphins}{\textcolor{dolphinsColor}{\textbf{Dolphins}}}
\newcommand{\omnidrive}{\textcolor{omnidriveColor}{\textbf{OmniDrive} (Omni-L)}}
\newcommand{\leapvad}{\textcolor{leapvadColor}{\textbf{LeapVAD}}}

\definecolor{success}{HTML}{2D7C32}
\definecolor{failure}{HTML}{C62727}

\definecolor{promptLightGray}{HTML}{F5F5F5}
\definecolor{promptBorderGray}{HTML}{B0B0B0}

\newcommand{\cmark}{\ding{51}}
\newcommand{\xmark}{\ding{55}}

\lstdefinestyle{promptstyle}{
  basicstyle=\ttfamily\scriptsize,      
  breaklines=true,                 
  breakindent=8pt,                 
  breakatwhitespace=false,
  backgroundcolor=\color{promptLightGray},
  frame=single,                    
  framerule=0.8pt,                 
  rulecolor=\color{promptBorderGray},    
  showspaces=false,
  showstringspaces=false,
  showtabs=false,
  xleftmargin=0.8pt,               
  xrightmargin=0.8pt,              
  resetmargins=true,               
  breakautoindent=false,           
  columns=flexible,                
  keepspaces=true,                 
}

\begin{document}

\title{Comparative Analysis of Patch Attack on VLM-Based Autonomous Driving Architectures}


\author{%
  \protect\parbox{\textwidth}{%
    \centering
    David Fernandez, Pedram MohajerAnsari, Amir Salarpour, Long Cheng, Abolfazl Razi, Mert D. Pes\'{e}%
  }%
  \thanks{David Fernandez, Pedram MohajerAnsari, Amir Salarpour, Long Cheng, Abolfazl Razi, and Mert D. Pes\'{e} are with the School of Computing, Clemson University, Clemson, SC, USA
    \{\texttt{dferna3}, \texttt{pmohaje}, \texttt{asalarp}, \texttt{lcheng2}, \texttt{arazi}, \texttt{mpese}\}@clemson.edu}
}

\maketitle

\begin{abstract}
Vision-language models are emerging for autonomous driving, yet their robustness to physical adversarial attacks remains unexplored. This paper presents a systematic framework for comparative adversarial evaluation across three VLM architectures: Dolphins, OmniDrive (Omni-L), and LeapVAD. Using black-box optimization with semantic homogenization for fair comparison, we evaluate physically realizable patch attacks in CARLA simulation. Results reveal severe vulnerabilities across all architectures, sustained multi-frame failures, and critical object detection degradation. Our analysis exposes distinct architectural vulnerability patterns, demonstrating that current VLM designs inadequately address adversarial threats in safety-critical autonomous driving applications.
\end{abstract}


\section{Introduction}
\label{sec:1-introduction}


Vision-language models (VLMs) are emerging in autonomous driving (AD), integrating visual perception with language-based reasoning to create interpretable, end-to-end decision-making systems~\cite{VLMsurvey}. Unlike traditional modular pipelines that separate perception, prediction, and planning, VLMs leverage large language models (LLMs), enabling them to handle complex driving scenarios through natural language (NL) understanding. Recent systems \cite{dolphins, drivegpt, lmdrive} show that VLMs can generate human-interpretable driving decisions and generalize to previously unseen scenarios.

Despite their promise, the robustness of VLM-based driving systems to physical adversarial attacks remains unclear. Physical adversarial patches pose a significant threat to safety-critical systems~\cite{eykholt2018robust}. While attacks on traditional vision models are well-studied~\cite{adversarialPatchBrown}, how they affect end-to-end VLM driving models presents a unique problem. The connection between vision and language creates a complex target, as attacks can corrupt visual data, distort scene understanding, and ultimately cause unsafe driving actions. This robustness question is further complicated by the proliferation of VLM architectures for AD.  Recent surveys~\cite{vlmList} identify over 20 distinct architectures with varying sensor modalities, vision encoders, language models, and fusion mechanisms. Many systems lack publicly available implementations (e.g., DriveGPT4~\cite{drivegpt}, VLP~\cite{pan2024vlp}, DriVLMe~\cite{huang2024drivlme}), preventing reproducible analysis. Others incorporate additional sensors such as LiDAR (e.g., LMDrive~\cite{lmdrive}) that fundamentally alter the attack surface, making fair architectural comparisons impractical. Furthermore, these architectures generate a wide range of outputs. As a result, no framework exists to evaluate how well different VLM architectures hold up to adversarial attacks or to compare them systematically, which limits our ability to understand which design choices improve robustness.

\begin{figure*}[!t]
    \centering
    \includegraphics[width=0.8\textwidth]{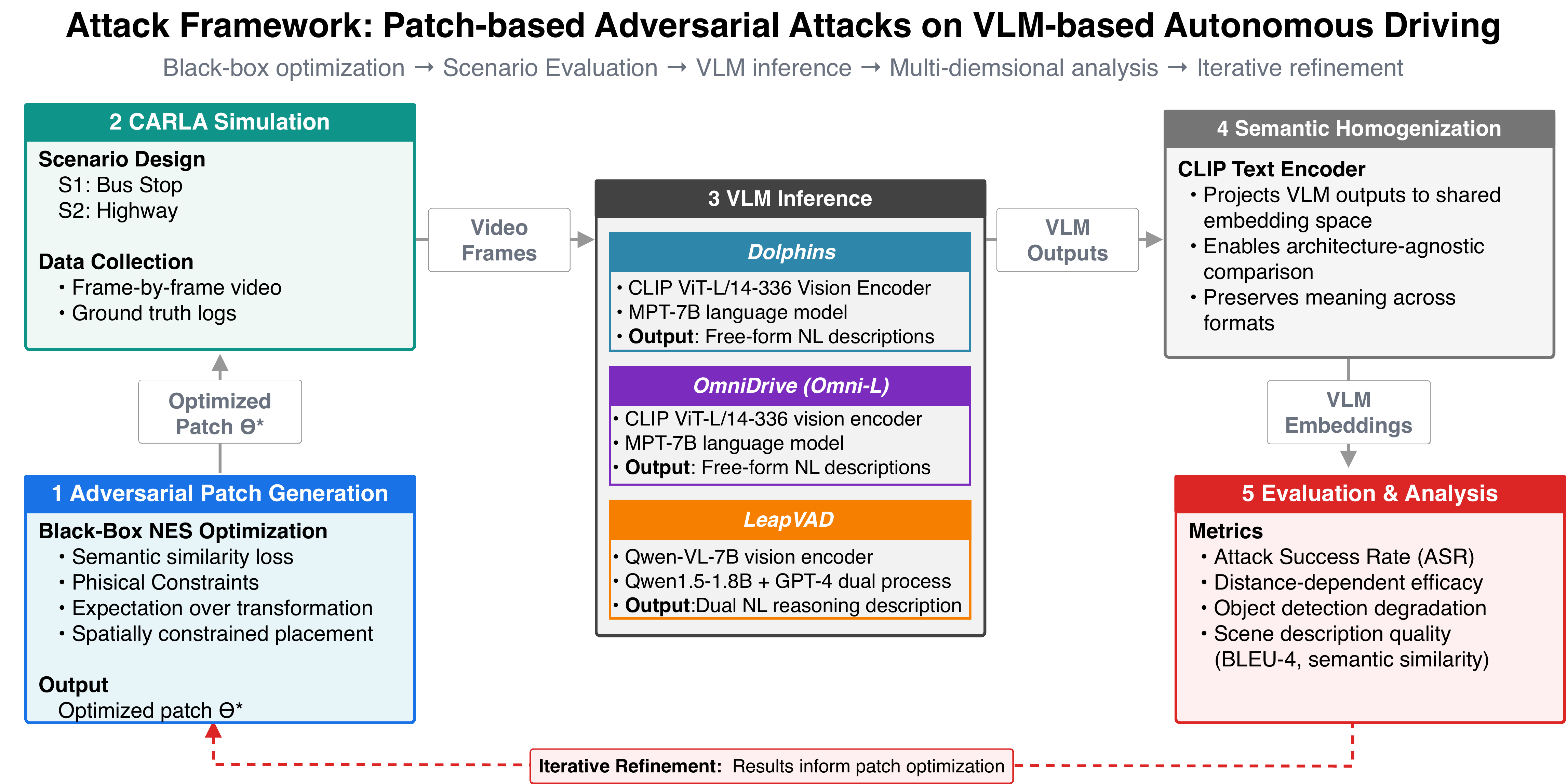}
    \caption{Overview of Attack Framework: (1) Black-box NES optimization generates adversarial patches; (2) CARLA scenarios capture frame video sequences; (3) VLM architectures process scenes: \dolphins, \omnidrive, and \leapvad; (4) Semantic homogenization layer projects all VLM outputs into a unified embedding space; (5) Multi-dimensional evaluation.}
    \label{fig:framework}
    \vspace{-10pt}
\end{figure*}

This paper introduces a systematic framework (\autoref{fig:framework}) for a comparative evaluation of adversarial robustness in VLM-based autonomous driving systems using the CARLA simulator~\cite{CARLA}. We address the challenge of model diversity by establishing rigorous selection criteria for comparable VLM architectures, designing a unified evaluation protocol that accounts for architectural differences while enabling a fair comparison, and developing a semantic-level attack and evaluation methodology that works across diverse output formats. Our framework employs black-box Natural Evolution Strategies (NES)~\cite{ilyas2018black} optimization with semantic similarity loss to generate physically realizable adversarial patches, ensuring attacks are architecture-agnostic and deployment-realistic.

We evaluate three representative VLM architectures: \dolphins~\cite{dolphins}, \omnidrive~\cite{wang2025omnidrive}, and \leapvad~\cite{ma2025leapvad}. These systems span the design space of vision-language integration while remaining comparable through a shared modality (camera-only), a shared task (closed-loop driving), and available open-source implementations. To ensure fairness across these architectures, which have different output formats, we introduce a semantic homogenization layer. This enables an architecture-agnostic comparison of attack effectiveness, perceptual degradation, and scene understanding corruption.

We ran closed-loop experiments in the CARLA simulator using two scenarios: one targeting crosswalk pedestrian detection suppression and another focused on highway steering manipulation. Our black-box NES~\cite{naturalEvolutionStratgies} optimizer generated physically constrained adversarial patches. These patches were placed on realistic advertising infrastructure (bus shelters and billboards) and used Expectation over Transformation (EoT)~\cite{ExpectationOverTransformation} to ensure they remained robust across various viewing conditions. For each VLM, we measured the attack success rate at different spatial distances, its temporal consistency (multi-frame persistence), its impact on object detection (failure to see critical objects), and the degradation of overall scene understanding (using BLEU-4 and semantic similarity).

Our evaluation reveals vulnerabilities across all tested architectures. The comparative evaluation also reveals distinct architectural vulnerability patterns. The adversarial patches achieved high overall attack success rates (ASRs) ranging from 73.5\% to 76.0\%, a 12-20x increase over baseline inappropriate action rates. These attacks proved both effective at critical distances (10-25 meters) and highly persistent, causing sustained failures for 6.2 to 7.8 consecutive frames. Finally, the patches caused worse scene understanding, with descriptions showing low BLEU-4 (0.18-0.31) and semantic similarity (0.49-0.67) scores when compared against their benign counterparts. This paper makes the following contributions:

\begin{itemize}
    \item A framework for architecture-agnostic adversarial evaluation which introduces a semantic homogenization layer to project heterogeneous VLM outputs into a unified embedding space to enable a fair, black-box comparison;

    \item A comprehensive evaluation methodology that uses multi-dimensional metrics to measure an attack's impact on action recommendations, object detection, scene understanding, spatial robustness, and temporal persistence;
    
    \item An empirical demonstration showing that adversarial patches can compromise VLM-based driving systems.
\end{itemize}

\section{Related Work}
\label{sec:2-related_work}

\noindent \textbf{VLMs for AD.} LLMs have recently been used for driving scenario analysis~\cite{fernandezWIP}, showing promise for more understandable explanations of driving scenes. Building on this, recent advances of LLMs have led to the development of VLMs for end-to-end AD. These architectures use diverse integration strategies: Dolphins employs cross-attention mechanisms, OmniDrive explores two architectural variants. Omni-Q uses Q-Former-based 3D perception alignment, while Omni-L leverages MLP projection. LeapVAD introduces a dual-process architecture that distinguishes between a fast Heuristic Process and a slow Analytic Process. While these systems show promising performance in normal scenarios~\cite{fernandez2025avoiding}, their robustness to adversarial attacks, especially how their different architectures affect vulnerability patterns, remains unexplored. Recent work has also explored lightweight LLMs for real-world deployment~\cite{slms, fernandez2026forensic}, though the security implications of such edge deployments remain understudied.

\noindent \textbf{Adversarial Attacks on Computer Vision.} Physical adversarial patches have proven to be a significant vulnerability for computer vision systems. Brown et al.~\cite{adversarialPatchBrown} showed that printed patterns could reliably fool image classifiers in the real world. Later research extended these attacks to object detectors~\cite{eykholt2018robust} and segmentation models~\cite{segmentation}, revealing that safety-critical perception components can be fooled by these carefully designed visual patterns. Eykholt et al.~\cite{eykholt2018robust} demonstrated attacks on traffic sign recognition using simple stickers, highlighting a direct threat to autonomous vehicle perception.

\noindent \textbf{Adversarial Robustness of Multimodal Models.} Recent work has started to examine adversarial vulnerabilities in vision-language models, showing that combining vision and language creates new attack surfaces not found in vision-only systems~\cite{mohajeransari2025attention}. Studies on CLIP~\cite{CLIP}, BLIP~\cite{li2022blip}, and LLaVA show that attacks can exploit these vision-language connections to cause specific wrong outputs, manipulating both visual encoders and cross-modal attention mechanisms. However, no prior work has systematically compared the adversarial robustness of different VLM architectures. It is still unknown if different integration methods have their own unique vulnerabilities. Our work addresses this gap by providing a comparative study that evaluates how effective adversarial patch attacks are at different distances, how long their effects persist across consecutive frames, and whether they maintain efficacy under varying real-world viewing conditions across three representative VLM architectures. More broadly, recent work in thermal calibration, scientific simulation, and explainable medical AI highlights why we should carefully evaluate robustness before deploying these models in safety-critical settings \cite{rajoli2023thermal,saberian2025hydroquantum,soltani2025explainable}.

\section{Threat Model}
\noindent \textbf{Adversary Capabilities.} Our threat model assumes an adversary has "black-box" access to the VLM-based driving system. This means the attacker can query the VLMs with images and receive model outputs, but cannot access internal model parameters, gradients, or training data. This assumption reflects a realistic deployment, where production systems typically protect their internal architecture and expose only query results. The attacker's goal is to cause specific unsafe driving actions by placing an adversarial patch in the environment.

\noindent \textbf{Attack Constraints.} We constrain adversarial patch placement to existing road infrastructure, such as advertisement panels, reflecting a practical attack vector rather than allowing arbitrary placement. We assume the attacker can identify high-traffic locations, optimize patches offline using limited queries, and physically deploy them by printing or compromising a display system. The attacker cannot modify the vehicle’s sensors, software, or other components, and must succeed using only passive visual manipulation.

\section{VLM Architectures}
\label{sec:5-VLM_architectures}

\noindent \textbf{VLM Selection.} To enable a fair comparative adversarial evaluation, we established a selection criteria from a comprehensive survey of end-to-end VLM-based driving systems~\cite{vlmList}. We required models to use only camera inputs to ensure the comparison focused purely on the vision-language architecture. We also required publicly available implementations for reproducibility (excluding systems like DriveGPT4, VLP, and DriVLMe), and architectural diversity in vision-language integration. From over 20 candidates, three systems satisfied all criteria while maximizing this architectural coverage: 

\noindent \dolphins~uses across-attention approach for vision-language integration in AD. The architecture uses a CLIP ViT-L/14-336 vision encoder to process 336x336 pixel RGB images and generates free-form natural language narrative.

\noindent \omnidrive~proposes two architectural variants for driving VLMs, Omni-Q (Q-Former-based) and Omni-L (MLP-projection-based). We evaluate Omni-L, which produces conversational text with embedded 3D spatial coordinates. 

\noindent \leapvad~has two components: a scene understanding module using Qwen-VL-7B that explicitly identifies "critical objects" (like pedestrians or traffic lights) and a dual-process decision module, which consists of a fast heuristic process for real-time decisions and a slow analytic Process for complex reasoning. 

To enable an architecture-agnostic comparison, we introduce a semantic homogenization layer. This layer projects all VLM outputs into a unified embedding space using a frozen CLIP text encoder. This enables our black-box NES optimizer to use a unified semantic similarity loss function that works equivalently across all architectures, capturing both action-level corruption (a change in the recommended action) and reasoning-level corruption (a degradation in the semantic justification) while respecting their architectural differences.

\section{Methodology}
\label{sec:6-Methodology}

\subsection{Research Questions}
Our comparative evaluation analyzes three key dimensions of adversarial vulnerability in VLM architectures. \textbf{RQ1 (Spatial and Temporal Robustness):} To what extent do adversarial patches maintain attack effectiveness across realistic vehicle approach distances, and do attacks cause sustained multi-frame failures or intermittent single-frame errors that temporal filtering might mitigate? \textbf{RQ2 (Perceptual-Behavioral Coupling):} Do patches that corrupt action recommendations also degrade detection of safety-critical objects (pedestrians, barriers), or can patches induce unsafe actions while preserving object detection through reasoning corruption? \textbf{RQ3 (Scope of Corruption)}: Do adversarial patches cause localized action corruption or holistic scene understanding degradation? 

\subsection{Experimental Setup}
All experiments were conducted in CARLA 0.9.14 (Town04). We use a single forward facing RGB camera at 1920x1080 resolution to match standard autonomous vehicle configurations. While real-world deployment involves additional visual complexity, prior research shows that adversarial vulnerabilities identified in simulation reliably transfer to physical systems\cite{kong2020physgan, AdvExamplesGoodfellow}. Town04 provides sufficient scene variety to evaluate VLMs robustness under realistic conditions.

\begin{figure*}[t]
    \centering
    \begin{tabular}{@{}ccc@{}}
        \includegraphics[width=0.25\textwidth]{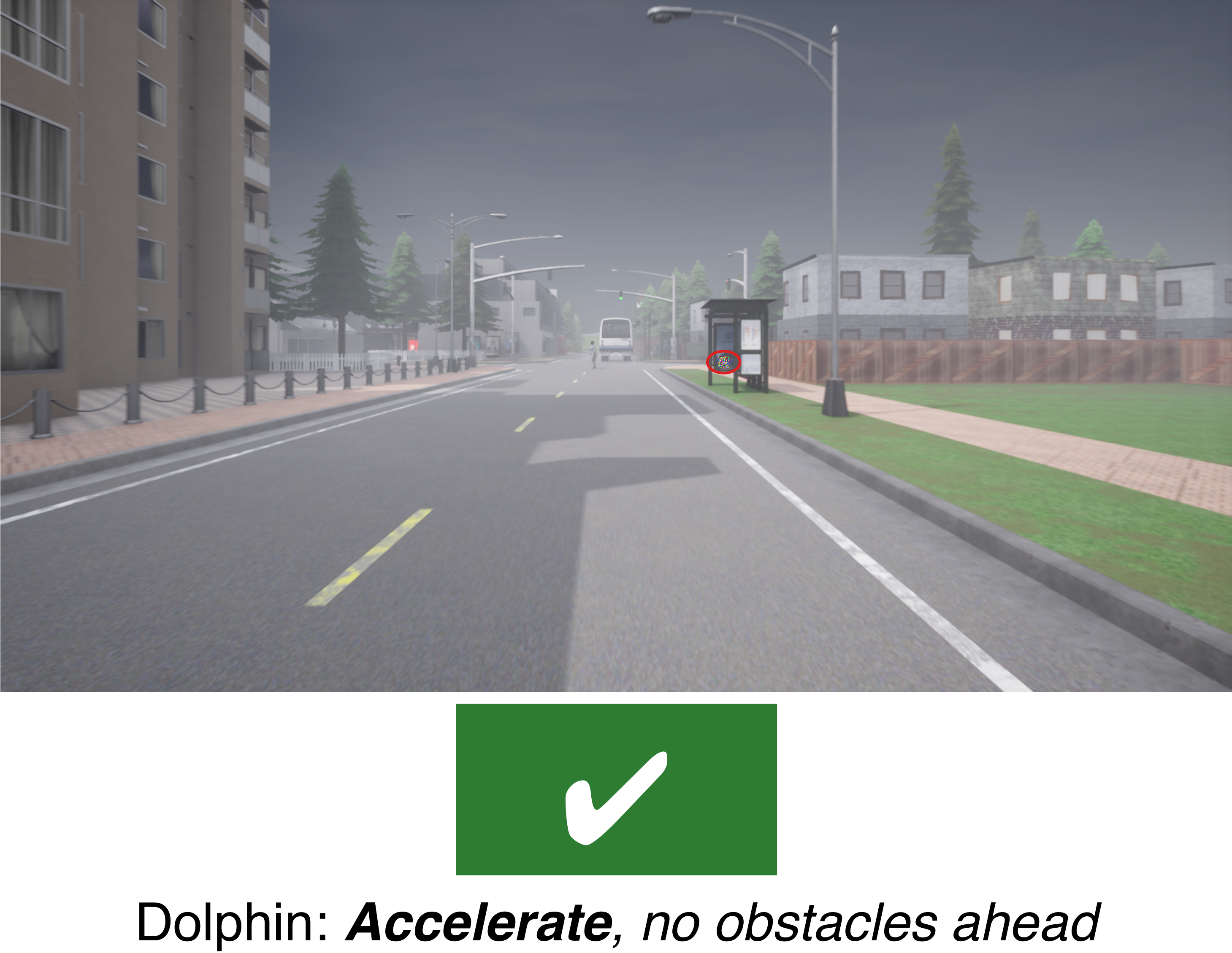} &
        \includegraphics[width=0.25\textwidth]{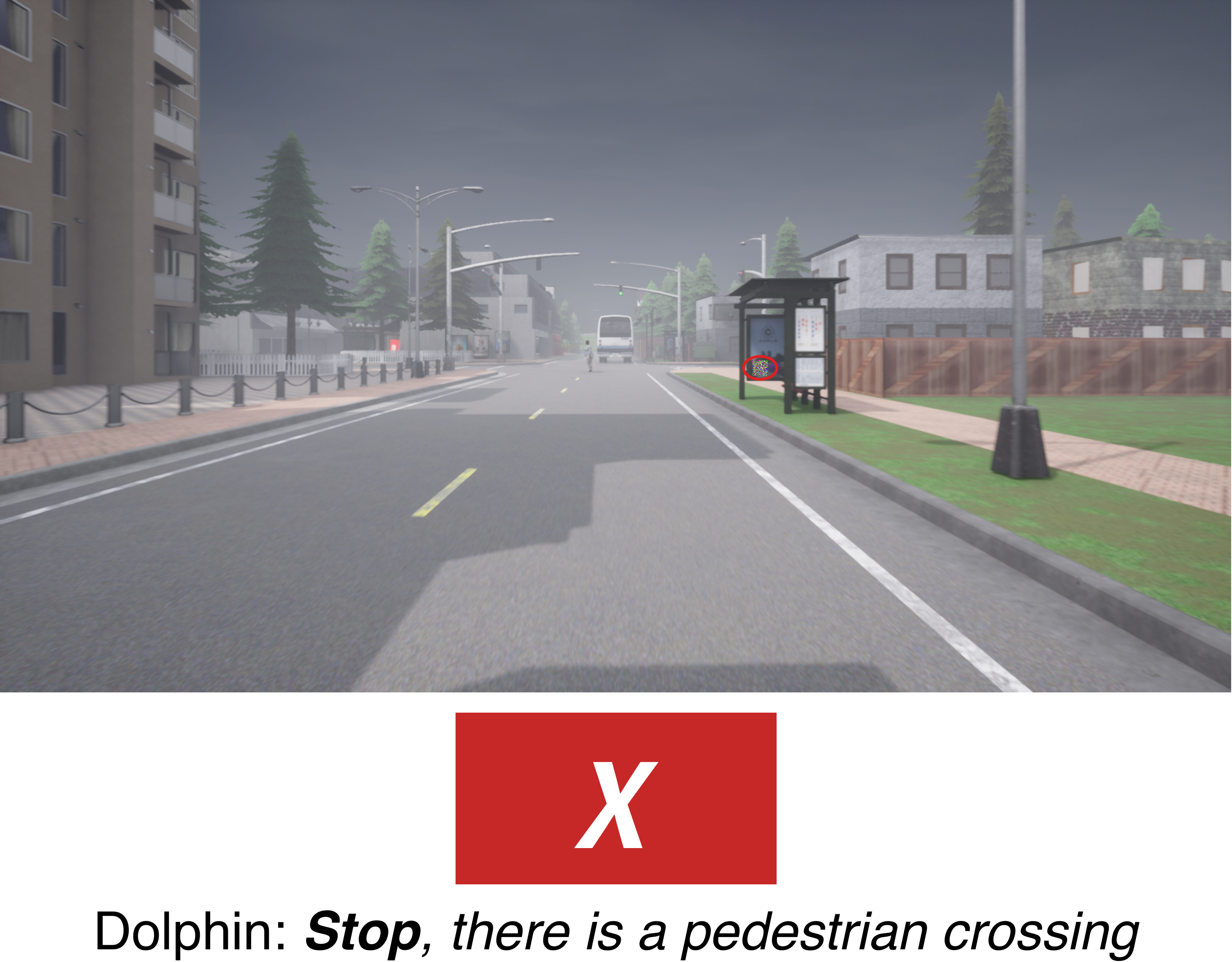} &
        \includegraphics[width=0.25\textwidth]{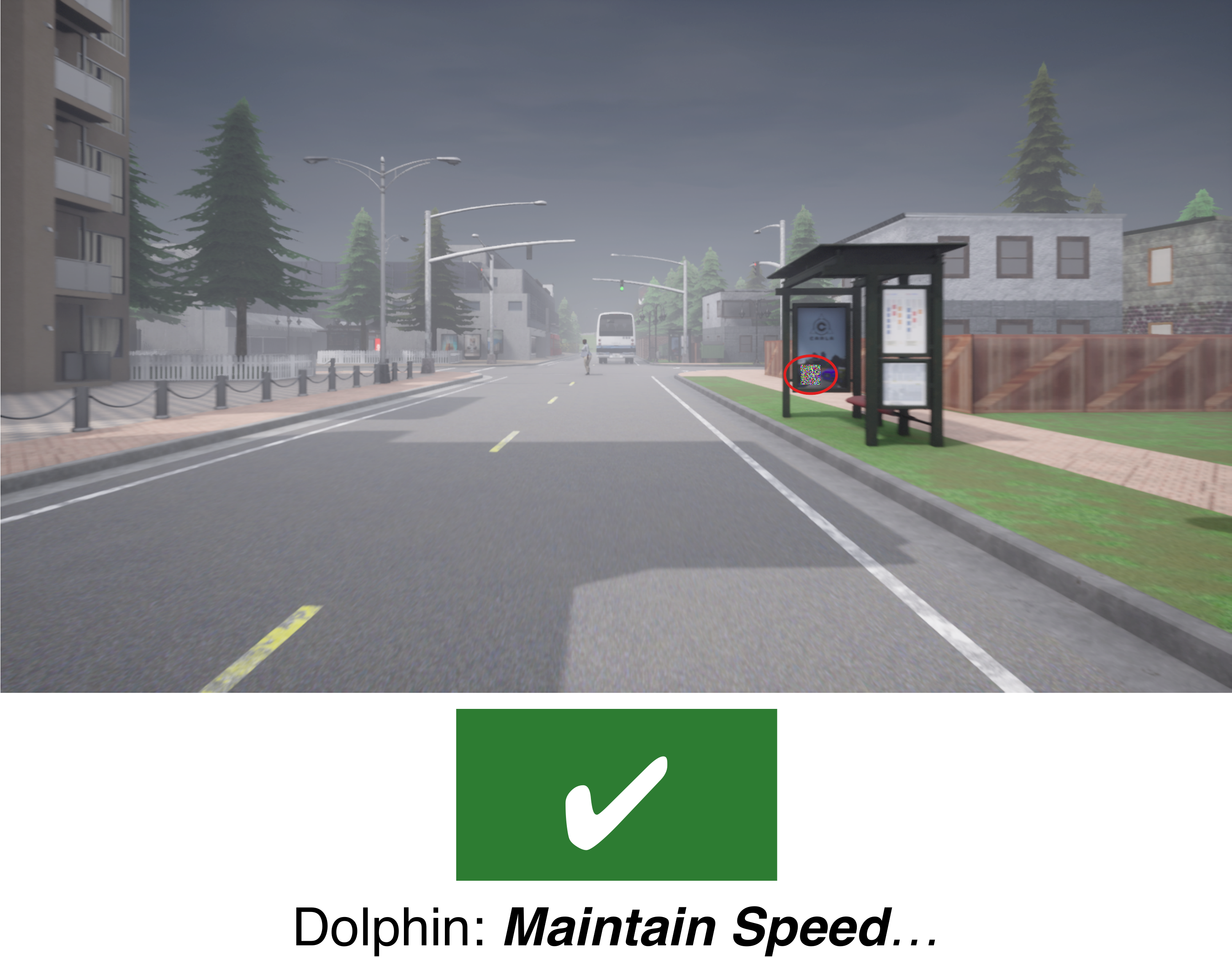} \\[2pt]
        
        \includegraphics[width=0.25\textwidth]{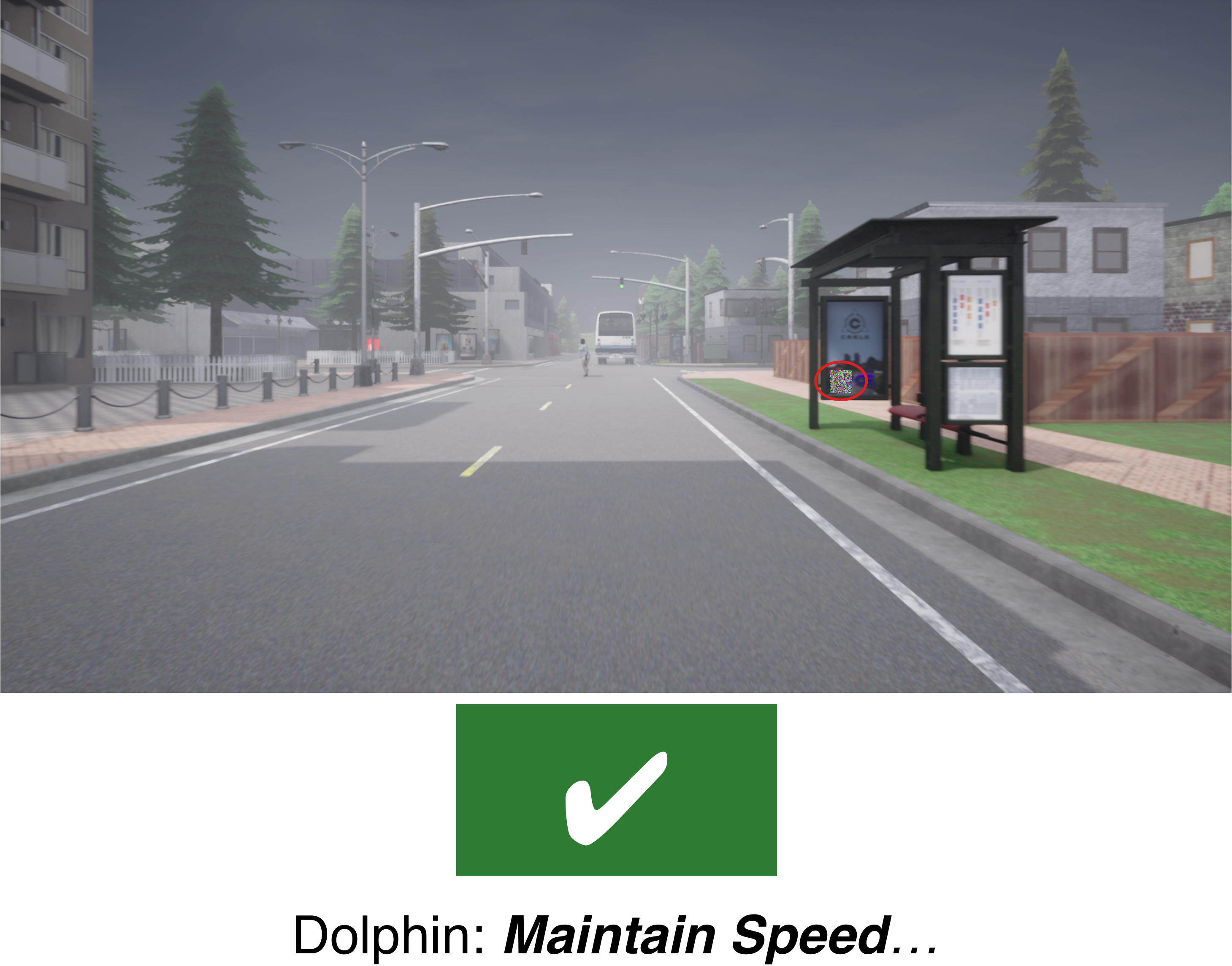} &
        \includegraphics[width=0.25\textwidth]{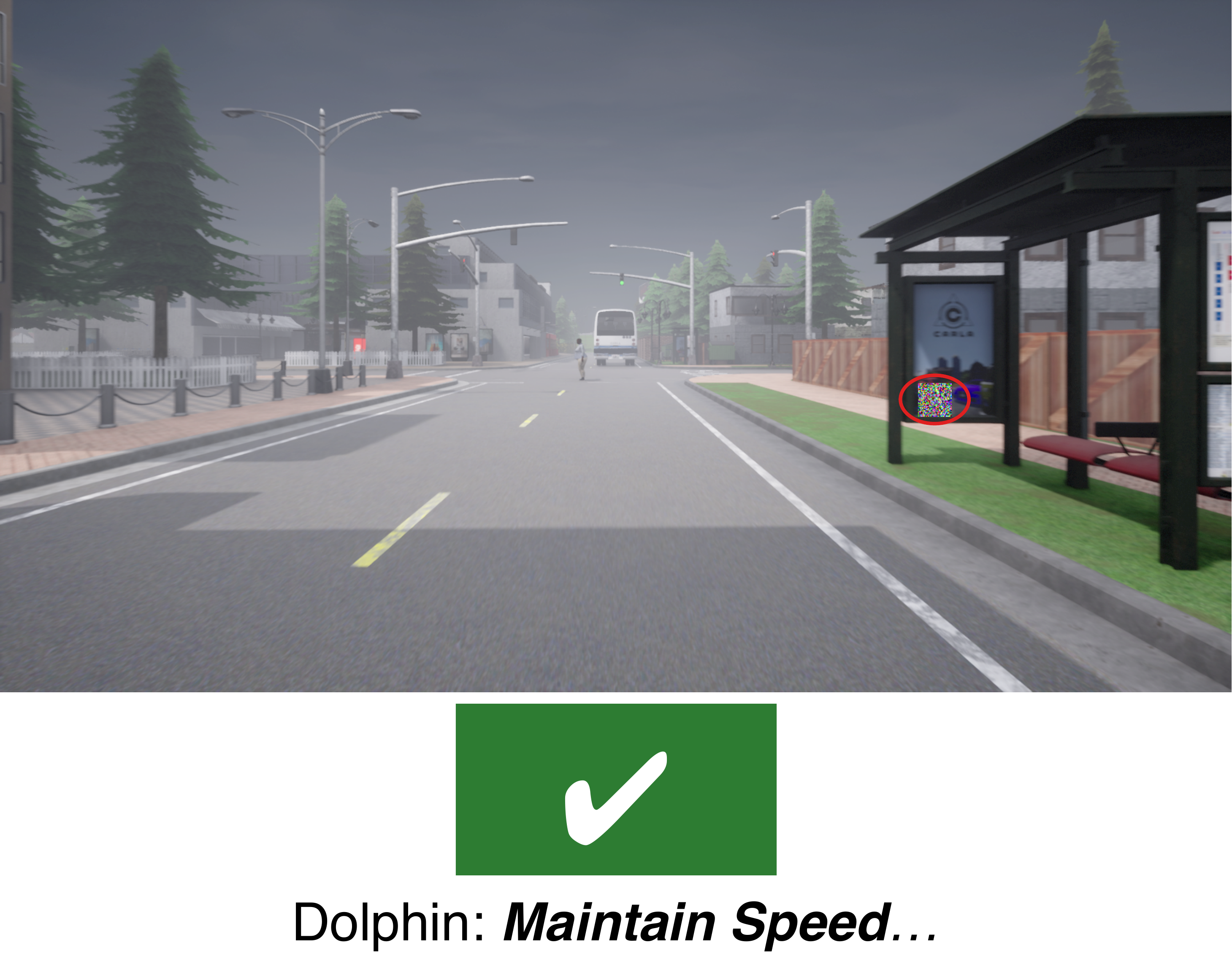} &
        \includegraphics[width=0.25\textwidth]{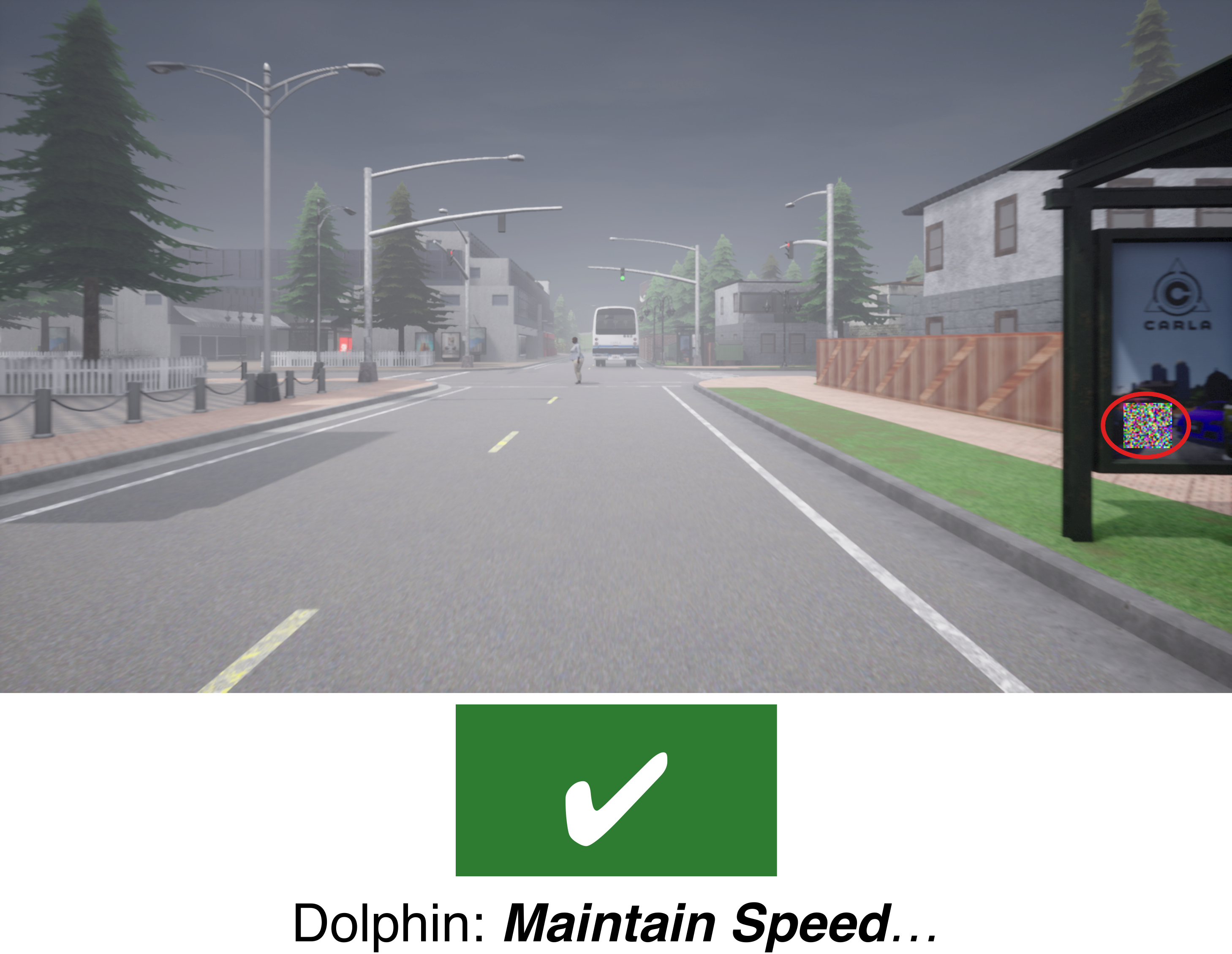} 
    \end{tabular}
    \caption{Scenario 1:Bus Shelter Crosswalk Attack temporal sequence. The attack scenario demonstrates how the adversarial patch suppresses pedestrian detection as the ego vehicle approaches bus ad shelter.}
    \label{fig:gridScenario1}
    \vspace{-10pt}
\end{figure*}


\subsection{Scenario Design}
We design two complementary scenarios that target different vulnerabilites in the VLMs perception-reasoning-action pipeline, enabling systematic evaluation of perceptual versus reasoning corruption across architectures.

\noindent \textbf{Scenario 1: Bus Shelter Crosswalk Attack.} As shown in~\autoref{fig:gridScenario1}, the first scenario is an urban intersection. The ego-vehicle approaches a crosswalk at 30 km/h as a pedestrian crosses its path. The scene includes a bus shelter with an advertisement panel and a bus on the opposite side of the road for realistic context. The attack involves placing a $512 \times 512$ pixel adversarial patch on the bus shelter's ad panel. From the vehicle's view, this patch becomes visible at about 30 meters and covers about 5-7\% of the camera's width at the critical decision point (10 meters from the crosswalk).

\noindent \textbf{Scenario 2: Highway Billboard Steering Attack.} The second scenario an ego-vehicle that travels at 85 km/h in the right lane of a highway. A large roadside billboard is near the right lane. Critically, a concrete barrier runs along the right side, meaning any rightward turn would be dangerous. The adversarial patch here is much larger, measuring $1024 \times 512$ pixels. It becomes visible from about 80 meters. Unlike the first scenario, this attack is optimized to corrupt the VLM's action, not its perception. The goal is to make the VLM recommend a "turn right" action (toward the barrier) instead of the safe "maintain speed" or "accelerate" actions.


%
%

\subsection{Adversarial Patch Generation}
We generate adversarial patches using NES, a gradient-free black-box optimization method for attacking VLMs without internal model access. NES estimates an objective function's gradient by strategically sampling the parameter space. In each iteration, it creates a "population" of new patches by adding small random noise to the current best patch. It then evaluates each new patch by querying the VLM and updates the main patch based on which variations were most successful.

\noindent \textbf{Optimization Process.} The patch is initialized with random Gaussian noise sampled from a standard normal distribution and scaled to the valid RGB range $[0, 255]$. At iteration $t$, the algorithm generates $N$ directional perturbations $\{\epsilon_i\}_{i=1}^{N}$ by sampling from $\mathcal{N}(0, \sigma^2 I)$, where $\sigma$ controls the exploration radius. For each perturbation direction, two candidate patches are evaluated: $\theta_t + \sigma \epsilon_i$ and $\theta_t - \sigma \epsilon_i$, where $\theta_t$ represents the current patch parameters. The objective function is computed for each candidate by placing it at a constrained location within the target image and querying Dolphin VLM for its action recommendation. The gradient estimate is then computed as:

\begin{equation}
\nabla_\theta J \approx \frac{1}{N\sigma} \sum_{i=1}^{N} [J(\theta + \sigma \epsilon_i) - J(\theta - \sigma \epsilon_i)] \epsilon_i
\end{equation}

and the patch is updated according to $\theta_{t+1} = \theta_t - \alpha \nabla_\theta J$, where $\alpha$ is the learning rate. This process repeats for a fixed number of iterations or until convergence criteria are met.

We used $N = 20$ perturbation directions per iteration, noise standard deviation $\sigma = 0.1$, and learning rate $\alpha = 0.02$. Each patch was optimized for 150 iterations, resulting in approximately 6,000 model queries per patch ($20$ directions $\times$ $2$ evaluations $\times$ $150$ iterations). These hyperparameters were selected through preliminary experiments to balance optimization quality against computational budget.

\noindent \textbf{Loss Function Design.} To enable architecture-agnostic optimization across VLMs with different outputs, we employ a semantic similarity loss based on CLIP text embeddings:

\begin{equation}
L_{\text{semantic}} = 1 - \frac{\mathbf{e}_{\text{generated}} \cdot \mathbf{e}_{\text{target}}}{\|\mathbf{e}_{\text{generated}}\| \|\mathbf{e}_{\text{target}}\|}
\label{eq:semantic_loss}
\end{equation}

\begin{table*}[t]
\centering
\caption{Overall attack success rates across VLM architectures. Attack success is defined as the proportion of frames where the VLM recommends the target unsafe action rather than the appropriate safety-critical action.}
\label{tab:overall_attack_vlms}
\resizebox{0.85\textwidth}{!}{%
\begin{tabular}{@{}lccccc@{}}
\toprule
\textbf{Model} 
& \textbf{Crosswalk ASR (\%)} 
& \textbf{Highway ASR (\%)} 
& \textbf{Combined ASR (\%)} 
& \textbf{Baseline (\%)} 
& \textbf{Significance} \\
\midrule
Dolphins 
& $73.1 \pm 2.8$ 
& $79.2 \pm 3.1$ 
& $76.0 \pm 2.4$ 
& 3.8 
& $p < 0.001$ \\
OmniDrive (Omni\mbox{-}L) 
& $71.8 \pm 3.2$ 
& $75.6 \pm 2.9$ 
& $73.5 \pm 2.6$ 
& 5.1 
& $p < 0.001$ \\
LeapVAD 
& $68.4 \pm 2.5$ 
& $81.7 \pm 2.7$ 
& $75.0 \pm 3.1$ 
& 6.3 
& $p < 0.001$ \\
\bottomrule
\end{tabular}%
}
\vspace{-10pt}
\end{table*}

where $\mathbf{e}_{\text{generated}}$ and $\mathbf{e}_{\text{target}}$ are CLIP text embeddings of the VLM's output and the desired target response, respectively. The term $\frac{\mathbf{e}_{\text{generated}} \cdot \mathbf{e}_{\text{target}}}{\|\mathbf{e}_{\text{generated}}\| \|\mathbf{e}_{\text{target}}\|}$ represents the cosine similarity between the two embedding vectors, which measures the angular alignment of the embeddings in the high-dimensional CLIP text space. Cosine similarity ranges from $-1$ (completely opposite meaning) to $+1$ (identical meaning), yielding a loss in [0,2] that provides smooth gradients for optimization.

For the crosswalk scenario, the target response was \textit{"The driver should accelerate and continue forward because the road is clear"}, while for the highway scenario, the target was \textit{"The driver should turn right to exit the highway"}. These target phrases encode both the desired action and contextual justification, encouraging patches to corrupt not only the action recommendation but also the underlying scene interpretation. 

\noindent \textbf{Physical Realizability Constraints.} To ensure that optimized patches could be printed and deployed in real-world scenarios, we enforced several constraints during optimization. All RGB pixel values were hard-clipped to range $[0, 255]$ at each iteration, guaranteeing compatibility with standard color printers and displays. Additionally, we applied weak total variation (TV) regularization with coefficient $\lambda_{\text{TV}} = 0.001$ to encourage spatial smoothness and discourage extreme high-frequency patterns that would be difficult to reproduce on physical substrates or could be easily detected by preprocessing filters, penalizing large differences between adjacent pixels without constraining the optimization's ability to find effective attack patterns. TV regularization is computed as:

\begin{equation}
\text{TV}(\theta) = \sum_{i,j} \left[ (\theta_{i+1,j} - \theta_{i,j})^2 + (\theta_{i,j+1} - \theta_{i,j})^2 \right]^{1/2}
\end{equation}

Patch dimensions were based on the physical targets: $512 \times 512$ (1 sq. meter) for the Scenario 1 and $1024 \times 512$ pixels (2m x 1m) for the Scenario 2. These sizes were selected to be visible at critical distances while remaining plausible as legitimate advertisements. This approach increases our threat model's realism by simulating an attacker who compromises existing advertising infrastructure, an attack vector that does not require introducing novel objects into the environment.

\noindent \textbf{Expectation Over Transformation (EoT)}. To enhance patch robustness, we incorporated EoT into the optimization. At each iteration, we applied random transformations sampled from realistic distributions to the patched image before querying the VLMs. This optimizes the patch to remain effective across various viewing conditions, not just a single static configuration. These transformations included spatial jittering (random translations of $\pm 5$ pixels), brightness adjustment (multiplicative factors in range $[0.9, 1.1]$), and contrast variation (additive shifts in range $[-0.05, 0.05]$). The loss function was then computed as the expectation across $K=5$ independently transformed samples per candidate patch:

\begin{equation}
\mathbb{E}_{T \sim \mathcal{T}}[L(\theta, T)] \approx \frac{1}{K} \sum_{k=1}^{K} L(\theta, T_k)
\end{equation}

where $\mathcal{T}$ represents the distribution of transformations and $T_k$ are sampled transformation instances. This EoT approach ensures that optimized patches remain effective when viewing conditions vary slightly from the exact optimization scenario.

\subsection{Evaluation Metrics}

\noindent \textbf{Attack Success Rate (ASR).} We define frame-wise ASR as the proportion of frames where the VLM recommends the target unsafe action (e.g., "accelerate" toward pedestrian, "turn right" toward barrier). For each scenario, we extract 8-12 frames per trial at 0.5-second intervals as the ego vehicle approaches adversarial infrastructure from the moment the patch first becomes visible in the camera frame until the vehicle passes the patch location. ASR is computed as:

\begin{equation}
\text{ASR}_{\text{frame}} = \frac{1}{N \cdot F} \sum_{i=1}^{N} \sum_{j=1}^{F_i} \mathds{1}[\text{action}_{i,j} = \text{target}]
\end{equation}

where $N$ is the number of trials, $F_i$ is the number of frames extracted from trial $i$, and $\mathbb{1}[\cdot]$ is the indicator function. Statistical significance is assessed using generalized estimating equations (GEE) to account for within-trial correlation of frames, comparing adversarial ASRs against baseline inappropriate action rates with $ p < 0.05 $ threshold.

\begin{figure*}[t]
\centering

\begin{minipage}{0.78\textwidth}
    \centering \small
    \setlength{\tabcolsep}{1em}
    \renewcommand{\arraystretch}{1.2}
    \begin{tabular}{|cccccc|}
        \hline
        \raisebox{0.2ex}{\tikz\draw[dolphinsColor,fill=dolphinsColor] (0,0) circle (2pt);} & \dolphins &
        \raisebox{0.2ex}{\tikz\draw[omnidriveColor,fill=omnidriveColor] (-2pt,-2pt) rectangle (2pt,2pt);} & \textcolor{omnidriveColor}{\textbf{Omni-L}} &
        \raisebox{0.2ex}{\tikz\draw[leapvadColor,fill=leapvadColor] (0,2.3pt) -- (-2.3pt,-2.3pt) -- (2.3pt,-2.3pt) -- cycle;} & \leapvad \\
        \hline
    \end{tabular}
\end{minipage}

\begin{tikzpicture}
\begin{axis}[
    width=0.45\textwidth,
    height=0.3\textwidth,
    xlabel={Distance to Patch (meters)},
    ylabel={Attack Success Rate (\%)},
    title={Scenario 1: Crosswalk Attack},
    xmin=0, xmax=42,
    ymin=20, ymax=105,
    xtick={0,5,10,15,20,25,30,35,40},
    ytick={20,30,40,50,60,70,80,90,100},
    grid=both,
    grid style={line width=0.1pt, draw=gray!30},
    major grid style={line width=0.2pt, draw=gray!40},
    xlabel style={font=\bfseries\small},
    ylabel style={font=\bfseries\small},
    title style={font=\bfseries\small},
    tick label style={font=\small}
]

\fill[red!10, opacity=0.3] (axis cs:10,20) rectangle (axis cs:25,105);

\addplot[
    color=dolphinsColor,
    mark=o,
    line width=2pt,
    mark size=2.5pt,
    mark options={solid, fill=dolphinsColor}
] coordinates {
    (3, 50.0) (7.5, 70.0) (15, 88.9) (25, 75.0) (35, 37.5)
};

\addplot[
    color=omnidriveColor,
    mark=square,
    line width=2pt,
    mark size=2.5pt,
    mark options={solid, fill=omnidriveColor}
] coordinates {
    (3, 55.0) (7.5, 75.0) (15, 91.7) (25, 82.5) (35, 45.0)
};

\addplot[
    color=leapvadColor,
    mark=triangle,
    line width=2pt,
    mark size=3pt,
    mark options={solid, fill=leapvadColor}
] coordinates {
    (3, 45.0) (7.5, 62.5) (15, 80.0) (25, 68.3) (35, 35.0)
};

\addplot[
    color=gray,
    line width=1pt,
    dotted,
    forget plot
] coordinates {(0, 50) (42, 50)};

\node[anchor=east, color=gray, font=\scriptsize] at (axis cs:16,46) {50\% threshold};

\end{axis}
\end{tikzpicture}
\begin{tikzpicture}
\begin{axis}[
    width=0.45\textwidth,
    height=0.30\textwidth,
    xlabel={Distance to Patch (meters)},
    ylabel={Attack Success Rate (\%)},
    title={Scenario 2: Highway Attack},
    xmin=0, xmax=42,
    ymin=20, ymax=105,
    xtick={0,5,10,15,20,25,30,35,40},
    ytick={20,30,40,50,60,70,80,90,100},
    grid=both,
    grid style={line width=0.1pt, draw=gray!30},
    major grid style={line width=0.2pt, draw=gray!40},
    xlabel style={font=\bfseries\small},
    ylabel style={font=\bfseries\small},
    title style={font=\bfseries\small},
    tick label style={font=\small}
]

\fill[red!10, opacity=0.3] (axis cs:10,20) rectangle (axis cs:25,105);

\addplot[
    color=dolphinsColor,
    mark=o,
    line width=2pt,
    mark size=2.5pt,
    mark options={solid, fill=dolphinsColor}
] coordinates {
    (3, 75.0) (7.5, 80.0) (15, 87.5) (25, 81.8) (35, 42.9)
};

\addplot[
    color=omnidriveColor,
    mark=square,
    line width=2pt,
    mark size=2.5pt,
    mark options={solid, fill=omnidriveColor}
] coordinates {
    (3, 78.0) (7.5, 84.5) (15, 90.6) (25, 85.0) (35, 48.5)
};

\addplot[
    color=leapvadColor,
    mark=triangle,
    line width=2pt,
    mark size=3pt,
    mark options={solid, fill=leapvadColor}
] coordinates {
    (3, 68.0) (7.5, 72.5) (15, 82.5) (25, 73.3) (35, 38.0)
};

\addplot[
    color=gray,
    line width=1pt,
    dotted,
    forget plot
] coordinates {(0, 50) (42, 50)};

\node[anchor=east, color=gray, font=\scriptsize] at (axis cs:14,46) {50\% threshold};

\end{axis}
\end{tikzpicture}

\caption{Distance-dependent attack success rates across VLM architectures for both scenarios. The shaded red region indicates the critical decision-making range. Omni-L (purple) exhibits highest vulnerability across distances, while LeapVAD (orange) shows superior robustness, particularly at close ranges where its explicit critical object attention provides maximal benefit.}
\label{fig:distance_comparative}
\vspace{-10pt}
\end{figure*}
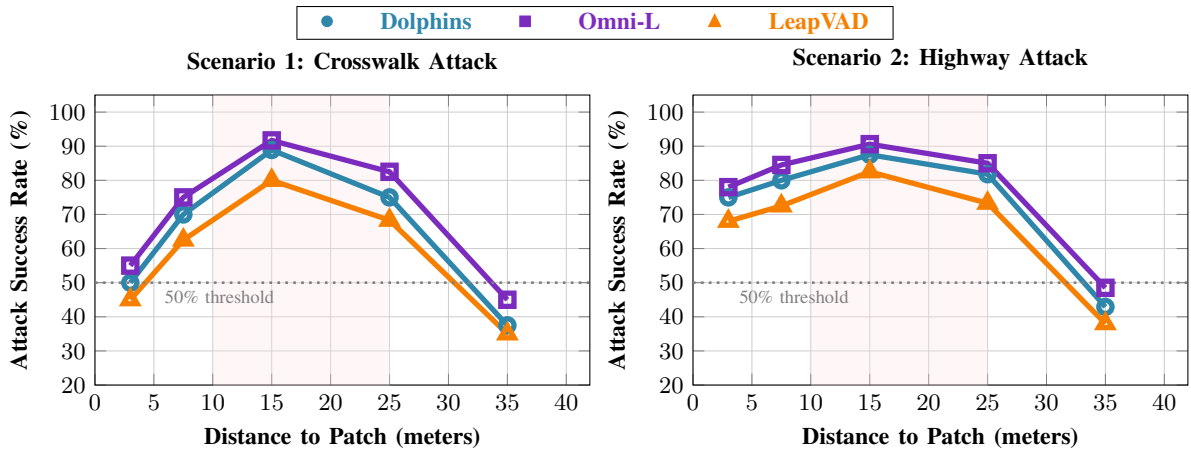

\noindent \textbf{Temporal Consistency and Attack Persistence.} To quantify attack stability across consecutive frames, we examine the sequence of per-frame attack outcomes (success or failure) and measure the length of consecutive successful attack frames. We define temporal persistence as:

\begin{equation}
\text{Persistence} = \mathbb{E}\left[\max_{k} \left\{ \ell : \prod_{j=k}^{k+\ell-1} \mathds{1}[\text{success}_j] = 1 \right\} \right]
\end{equation}

High persistence values (approaching total frame count) indicate sustained failures that would reliably mislead autonomous vehicles, while low persistence suggests intermittent successes that might be mitigated through temporal filtering.

\noindent \textbf{Object Detection Rate.} We perform keyword-based entity extraction on VLM-generated scene descriptions, searching for scenario-specific critical objects: pedestrian-related terms ("pedestrian," "person," "walker," "crossing") for Scenario 1, and infrastructure terms ("barrier," "wall," "concrete," "guard rail") for Scenario 2. Detection rate is computed as:

\begin{equation}
\text{DR}_{\text{frame}} = \frac{1}{N \cdot F} \sum_{i=1}^{N} \sum_{j=1}^{F_i} \mathds{1}[\text{critical object detected}_{i,j}]
\label{eq:detection_rate}
\end{equation}

for both benign and adversarial conditions. Detection degradation quantifies the percentage point (pp) decrease in detection rate caused by adversarial patches: $\Delta \text{DR} = \text{DR}_{\text{benign}} - \text{DR}_{\text{adv}}$. Values approaching 100\% indicate that patches completely suppress safety-critical object detection across the approach trajectory, representing critical safety failures.

\noindent \textbf{Scene Description Quality Metrics.} For selected frames at key distances (10m, 20m, 30m), we compute BLEU-4 scores~\cite{bleu} and cosine similarity of Sentence-BERT~\cite{sbert} embeddings between frame-matched benign and adversarial descriptions. BLEU-4 measures 4-gram overlap (0=no overlap, 1=identical), capturing lexical similarity, while semantic similarity (using SBERT) measures meaning preservation in embedding space. Low scores indicate patches corrupt holistic scene understanding beyond isolated action errors.

\section{Evaluation and Analysis}
\label{sec:6-Evaluation}

\subsection{Overall Attack Performance}

~\autoref{tab:overall_attack_vlms} presents aggregate attack success rates across both scenarios for all three VLM architectures. Adversarial patches reveal severe vulnerabilities across all systems, with overall ASRs ranging from 73.5\% to 76.0\%, representing a 12-20x increase over baseline inappropriate action rates (3.8-6.3\%). However, architectural differences are apparent in the attack patterns. Dolphins exhibits the highest vulnerability in the crosswalk scenario (73.1\% ASR) but more intermediate highway vulnerability (79.2\%), suggesting its cross-attention mechanism is particularly susceptible to perceptual corruption attacks. OmniDrive (Omni-L) demonstrates the most consistent performance across scenarios (71.8\% crosswalk, 75.6\% highway), potentially reflecting the robustness benefits of its MLP projection bottleneck that limits information flow. LeapVAD shows an inverted vulnerability pattern compared to Dolphins. It had a lower crosswalk ASR (68.4\%) but the highest highway ASR (81.7\%). This indicates that its explicit critical object attention provides partial protection against pedestrian suppression attacks but concentrates vulnerability when reasoning about spatial infrastructure.

\subsection{Distance-Dependent Attack Efficacy}
Distance-dependent analysis (Figure~\ref{fig:distance_comparative}) shows distinct vulnerability profiles correlated with VLM architecture. All systems failed at extreme distances (30m+ or $<5$m) due to patch size or distortion. However, critical differences emerged in the safety-critical 10-25m decision range where the AV must commit to braking, steering, or acceleration actions.

OmniDrive (Omni-L) was consistently the most vulnerable, maintaining 82-91\% ASR for scenario 1 and 85-90\% for scenario 2. This elevated vulnerability profile reflects its MLP projection architecture. The fixed linear transformation from visual to language space provides uniform susceptibility, meaning it is vulnerable regardless of the patch's apparent size.

Dolphins showed intermediate, distance-dependent vulnerability. Its attack success peaked at medium distances (e.g., 88.9\% at 15m) before declining at closer ranges. This pattern reflects its cross-attention architecture, which is optimally corrupted at mid-range but may suppress the patch's effectiveness when it becomes too anomalous at close range.

LeapVAD was the most robust, with 8-16 pp lower ASR than others in the critical range. Its advantage increased at close distances (3-7.5m), where it maintained superior resistance (45-62.5\% ASR). As pedestrians or barriers get closer and occupy more of the frame, LeapVAD's dedicated module can more reliably detect them despite adversarial interference.

The highway scenario shows smaller differences in vulnerability compared to the crosswalk, with all models achieving 5 to 8 pp higher attack success. This likely reflects the larger patch size and the more stable, frontal viewing angle, which provide the adversarial patterns with greater visual influence.

\subsection{Temporal Consistency and Attack Persistence}



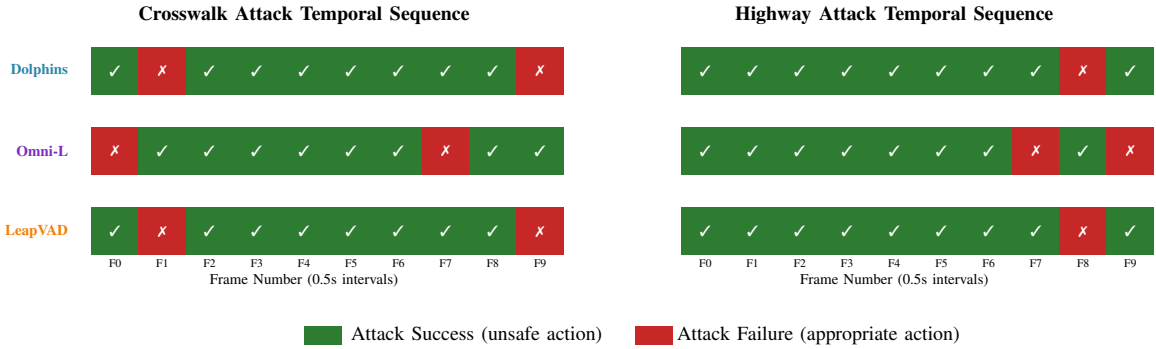
\begin{figure*}[t]
\centering
\resizebox{0.90\textwidth}{!}{%
\begin{tikzpicture}[
  every node/.style={font=\small, inner sep=0pt, outer sep=0pt}
]
\path[use as bounding box] (-2.0,-4.5) rectangle (24.0,3.5);

\def\xCrosswalk{0}
\def\xHighway{12.5}

\def\yD{1.5}
\def\yO{-0.2}
\def\yL{-1.9}

\node[font=\large\bfseries, anchor=south] at (\xCrosswalk+4.5, 3.0)
  {Crosswalk Attack Temporal Sequence};
\node[font=\large\bfseries, anchor=south] at (\xHighway+4.5, 3.0)
  {Highway Attack Temporal Sequence};

\node[font=\small\bfseries, anchor=east] at (-0.5, \yD+0.5) {\dolphins};
\node[font=\small\bfseries, anchor=east] at (-0.5, \yO+0.5) {\textcolor{omnidriveColor}{\textbf{Omni-L}}};
\node[font=\small\bfseries, anchor=east] at (-0.5, \yL+0.5) {\leapvad};

\newcommand{\DrawRow}[3]{%
  \foreach \i/\outcome in {#3} {%
    \pgfmathsetmacro{\xpos}{#1 + \i}
    \ifnum\outcome=1
      \fill[success] (\xpos, #2) rectangle +(1, 1);
      \node[white, font=\small\bfseries] at (\xpos+0.5, #2+0.5) {\cmark};
    \else
      \fill[failure] (\xpos, #2) rectangle +(1, 1);
      \node[white, font=\small\bfseries] at (\xpos+0.5, #2+0.5) {\xmark};
    \fi
  }%
}

\DrawRow{\xCrosswalk}{\yD}{0/1, 1/0, 2/1, 3/1, 4/1, 5/1, 6/1, 7/1, 8/1, 9/0}
\DrawRow{\xHighway}{\yD}{0/1, 1/1, 2/1, 3/1, 4/1, 5/1, 6/1, 7/1, 8/0, 9/1}

\DrawRow{\xCrosswalk}{\yO}{0/0, 1/1, 2/1, 3/1, 4/1, 5/1, 6/1, 7/0, 8/1, 9/1}
\DrawRow{\xHighway}{\yO}{0/1, 1/1, 2/1, 3/1, 4/1, 5/1, 6/1, 7/0, 8/1, 9/0}

\DrawRow{\xCrosswalk}{\yL}{0/1, 1/0, 2/1, 3/1, 4/1, 5/1, 6/1, 7/1, 8/1, 9/0}
\DrawRow{\xHighway}{\yL}{0/1, 1/1, 2/1, 3/1, 4/1, 5/1, 6/1, 7/1, 8/0, 9/1}

\foreach \i in {0,1,...,9} {
  \node[font=\scriptsize, anchor=north] at (\xCrosswalk+\i+0.5, \yL-0.1) {F\i};
}
\node[font=\small, anchor=north] at (\xCrosswalk+4.5, \yL-0.4)
  {Frame Number (0.5s intervals)};

\foreach \i in {0,1,...,9} {
  \node[font=\scriptsize, anchor=north] at (\xHighway+\i+0.5, \yL-0.1) {F\i};
}
\node[font=\small, anchor=north] at (\xHighway+4.5, \yL-0.4)
  {Frame Number (0.5s intervals)};

\begin{scope}[shift={(5,-3.8)}]
  \fill[success] (-0.5, 0) rectangle +(0.8, 0.4);
  \node[anchor=west, font=\large] at (0.5, 0.2) {Attack Success (unsafe action)};
  \fill[failure] (6.5, 0) rectangle +(0.8, 0.4);
  \node[anchor=west, font=\large] at (7.4, 0.2) {Attack Failure (appropriate action)};
\end{scope}

\end{tikzpicture}%
}
\caption{Temporal attack persistence across VLM architectures. Each row shows representative trials for crosswalk and highway scenarios. Models demonstrate similar temporal consistency patterns with LeapVAD showing slightly longer attack persistence (7.8$\pm$1.4 frames highway) compared to OmniDrive (6.9$\pm$1.3 frames).}
\label{fig:temporal_comparison}
\end{figure*}

\autoref{fig:temporal_comparison} shows the temporal persistence analysis. All architectures exhibit sustained multi-frame failures rather than intermittent single-frame errors, with average persistence ranging from 6.2 to 7.8 consecutive frames.

Dolphins shows high temporal consistency (80\% of trials achieve $\geq$5 consecutive frames), with particularly sustained failures in highway scenarios (7.4 frames). This indicates that once cross-attention mechanisms are corrupted by adversarial features, the failure persists across subsequent frames.

OmniDrive (Omni-L) exhibits slightly lower persistence (6.2-6.9 frames), potentially reflecting frame-to-frame independence from its stateless MLP projection. Unlike cross-attention that maintains temporal context, MLP projection processes each frame independently, occasionally breaking attack persistence. However, this provides minimal practical benefit since 3+ second failures remain catastrophic.

\begin{table*}[t]
\centering
\scriptsize
\caption{Detection rates and scene description quality metrics across VLM architectures. Detection degradation is reported in percentage points (pp). BLEU-4 and semantic similarity scores compare adversarial vs. benign descriptions.}
\label{tab:detection_and_quality}
\setlength{\tabcolsep}{3pt}
\resizebox{0.70\textwidth}{!}{%
\begin{tabular}{lcccccccccc}
\toprule
& \multicolumn{5}{c}{\textbf{Crosswalk (Pedestrian)}} 
& \multicolumn{5}{c}{\textbf{Highway (Barrier)}} \\
\cmidrule(lr){2-6} \cmidrule(lr){7-11}
\textbf{Model} 
& \textbf{Benign} & \textbf{Adv.} & \textbf{Degrad.} & \textbf{BLEU-4} & \textbf{Sem.}
& \textbf{Benign} & \textbf{Adv.} & \textbf{Degrad.} & \textbf{BLEU-4} & \textbf{Sem.} \\
& \textbf{DR (\%)} & \textbf{DR (\%)} & \textbf{(pp)} & & \textbf{Sim.}
& \textbf{DR (\%)} & \textbf{DR (\%)} & \textbf{(pp)} & & \textbf{Sim.} \\
\midrule
Dolphins 
& 92.3 & 21.2 & $-71.1$ & 0.18 & 0.49
& 88.5 & 47.9 & $-40.6$ & 0.24 & 0.59 \\
OmniDrive (Omni\mbox{-}L) 
& 89.7 & 34.8 & $-54.9$ & 0.22 & 0.54
& 91.2 & 58.3 & $-32.9$ & 0.28 & 0.63 \\
LeapVAD 
& 94.6 & 48.2 & $-46.4$ & 0.26 & 0.58
& 87.4 & 52.1 & $-35.3$ & 0.31 & 0.67 \\
\bottomrule
\end{tabular}%
}
\vspace{-10pt}
\end{table*}
LeapVAD demonstrates the highest persistence (7.1-7.8 frames, 100\% of trials $\geq$5 consecutive frames), particularly in highway scenarios. We attribute this to its memory bank-based few-shot prompting. These results indicate that temporal filtering defenses are ineffective against adversarial VLM attacks. Multi-frame consensus mechanisms requiring 3-5 frame agreement would fail, as attacks persist for 6-8 frames. 



\subsection{Object Detection Degradation}


As shown in~\autoref{tab:detection_and_quality}, during baseline VLM performance, all three architectures demonstrated strong scene understanding, achieving pedestrian detection rates of 89.7\% to 94.6\% and barrier detection rates of 87.4\% to 91.2\%. However, under adversarial conditions, all tested models exhibit severe detection degradation. Dolphins suffers pedestrian detection degradation (-71.1pp), indicating patches severely corrupt its CLIP vision encoder. The cross-attention mechanism becomes a vulnerability, as corrupted visual tokens propagate directly to the language model. Barrier detection degradation was lower (-40.6pp), suggesting larger objects are harder to suppress.


OmniDrive (Omni-L) shows intermediate degradation (-54.9pp pedestrian, -32.9pp barrier), with detection rates between Dolphins and LeapVAD. The MLP bottleneck seems to offer partial robustness by limiting feature propagation. 

LeapVAD exhibits the best, though still inadequate, detection robustness (-46.4pp pedestrian, -35.3pp barrier), due to its explicit critical object attention. This specialized training provides some resistance to perceptual attacks. However, partial detection (48.2\% under attack) did not translate to safe actions, highlighting the perception-behavior decoupling (RQ2).


\subsection{Scene Understanding Quality Degradation}

~\autoref{tab:detection_and_quality} presents scene description quality metrics comparing benign vs. adversarial conditions at key decision distances (10m, 20m, 30m). All architectures exhibit substantial corruption beyond isolated action errors. Low BLEU-4 scores (0.18-0.31) indicate minimal textual overlap between benign and adversarial descriptions, while semantic similarity scores (0.49-0.67) confirm a fundamental divergence in meaning. Qualitative analysis reveals that VLMs generate coherent but false descriptions. For example, Dolphins produces fluent narratives omitting pedestrians ('The road ahead is clear...'), Omni-L outputs structured JSON with hallucinated spatial configurations, and LeapVAD provides logical-sounding but incorrect reasoning. An architectural pattern also emerged. LeapVAD consistently achieved the highest quality scores (0.26 BLEU-4, 0.58 semantic similarity average), suggesting its dual-process architecture provides some semantic robustness. However, even these scores represent severe degradation.

\section{Conclusion}
\label{sec:7-Conclusion}

This work demonstrates critical vulnerabilities in VLM-based AD systems through a systematic comparative evaluation of physical adversarial patch attacks across three architectures. Our experiments reveal that patches placed on realistic advertising infrastructure reliably compromise all tested systems. These attacks remain effective at critical decision-making distances, cause sustained multi-frame failures, and produce perceptual degradation. The attacks corrupt scene understanding, with adversarial descriptions showing minimal overlap when compared to benign counterparts. Our comparative analysis exposes distinct architectural vulnerability patterns and fundamental robustness tradeoffs. Dolphins' cross-attention mechanism enables holistic perceptual corruption, yielding the highest detection degradation. In contrast, OmniDrive provides spatially consistent but limited robustness that restricts both adversarial propagation and adaptive recovery. LeapVAD shifts vulnerability from perception to reasoning through its explicit critical object attention. While CARLA provides photorealistic and validated sensor models, it cannot fully capture all real-world complexity. Empirical validation through physical adversarial displays in closed-course testing represents critical future work.


\section*{Acknowledgement}

This work was supported in part by a grant from The BMW
Group, and in part by Clemson University’s Virtual Prototyping of Autonomy Enabled Ground Systems (VIPR-GS), under Cooperative Agreement W56HZV-21-2-0001 with the US Army DEVCOM Ground Vehicle Systems Center (GVSC). DISTRIBUTION STATEMENT A. Approved for public release; distribution is unlimited. OPSEC \# 10193

\bibliographystyle{plain}
\bibliography{ref}

\end{document}